\documentclass[conference]{IEEEtran}
\IEEEoverridecommandlockouts
\usepackage{cite}
\usepackage{amsmath,amssymb,amsfonts}
\usepackage{algorithmic}
\usepackage{graphicx}
\usepackage{textcomp}
\usepackage{xcolor}
\def\BibTeX{{\rm B\kern-.05em{\sc i\kern-.025em b}\kern-.08em
    T\kern-.1667em\lower.7ex\hbox{E}\kern-.125emX}}
\begin{document}

\title{Automated Data Curation Using GPS \& NLP to Generate Instruction-Action Pairs for Autonomous Vehicle Vision-Language Navigation Datasets   \\
\thanks{}
}
\author
{\IEEEauthorblockN{Guillermo Roque, Erika Maquiling, Jose Giovanni Tapia Lopez, Ross Greer}
\IEEEauthorblockA{
\textit{Machine Intelligence, Interaction, and Imagination (Mi$^3$) Laboratory}\\
\textit{University of California, Merced}}
}
\maketitle
\begin{abstract}
Instruction-Action (IA) data pairs are valuable for training robotic systems, especially autonomous vehicles (AVs), but having humans manually annotate this data is costly and time-inefficient. This paper explores the potential of using mobile application Global Positioning System (GPS) references and Natural Language Processing (NLP) to automatically generate large volumes of IA commands and responses without having a human generate or retroactively tag the data. In our pilot data collection, by driving to various destinations and collecting voice instructions from GPS applications, we demonstrate a means to collect and categorize the diverse sets of instructions, further accompanied by video data to form complete vision-language-action triads. We provide details on our completely automated data collection prototype system, ADVLAT-Engine. We characterize collected GPS voice instructions into eight different classifications, highlighting the breadth of commands and referentialities available for curation from freely available mobile applications. Through research and exploration into the automation of IA data pairs using GPS references, the potential to increase the speed and volume at which high-quality IA datasets are created, while minimizing cost, can pave the way for robust vision-language-action (VLA) models to serve tasks in vision-language navigation (VLN) and human-interactive autonomous systems. 

\end{abstract}

\begin{IEEEkeywords}
Autonomous Vehicles, Natural Language Processing, Vision-Language Models, Vision-Language-Action Models, Vision-Language Navigation
\end{IEEEkeywords}

\section{Introduction}

\begin{figure}
    \centering
    \includegraphics[width=0.49\textwidth]{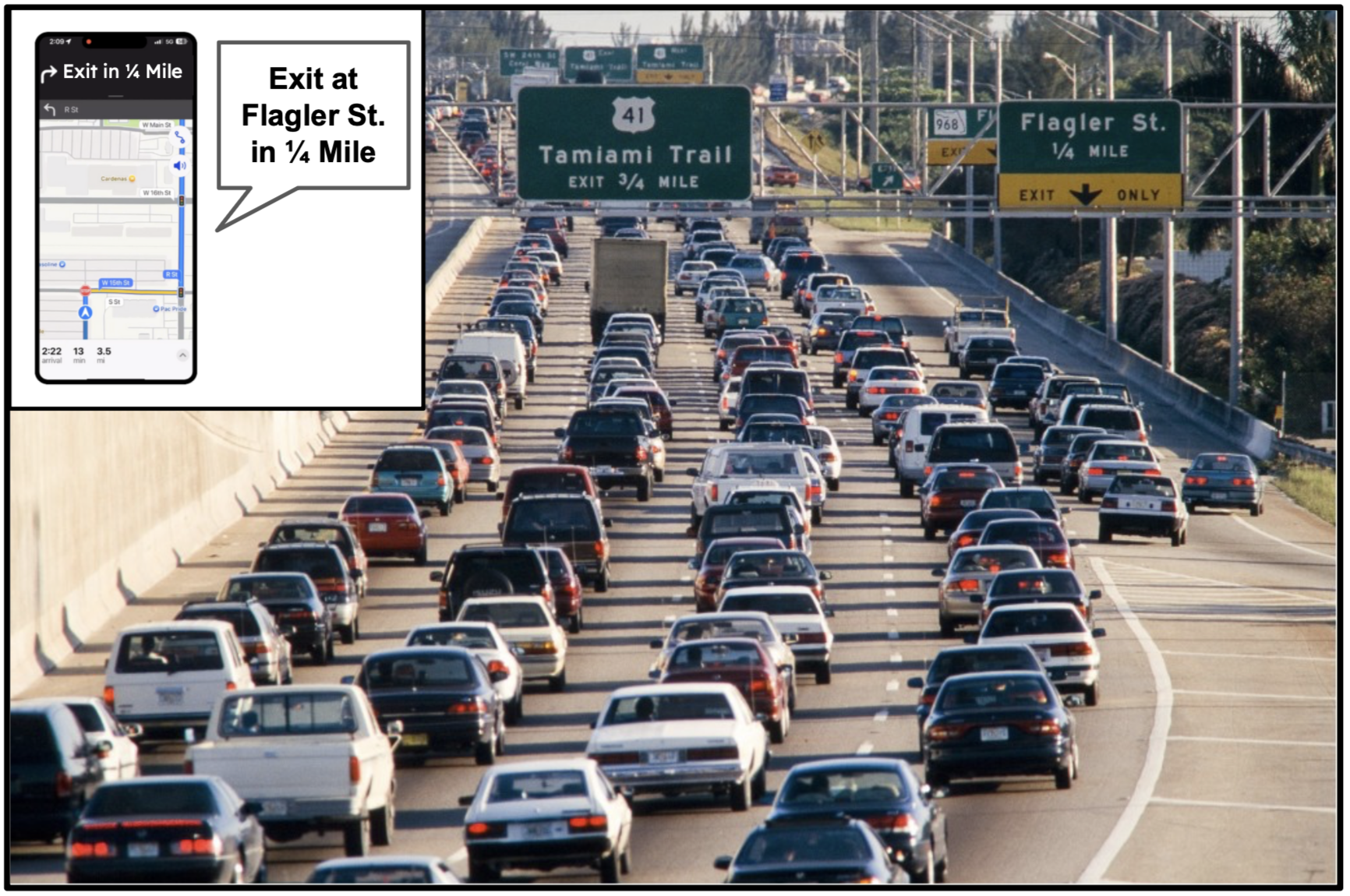} 
    \caption{The prevalent usage of the Global Positioning System (GPS) navigation service apps forms a plethora of valuable natural language instructions to drivers. In this research, we highlight the unexplored potential of this data for use in the autonomous driving field.}
    \label{fig:sample_intro}
\end{figure}

The ever-changing field of autonomous driving has led to an increased demand for Instruction-Action (IA) pair datasets \cite{shah2021ving}. There is strong incentive to generate these data pairs in great quantity to train models to assist and drive autonomous vehicles, or more broadly, robots, in semi-automated navigation and decision making 
\cite{shah2023lm, hao2020towards, gu2022vision}. In this research, we propose that an untapped source of such instruction-action pairs is available through standard GPS navigation apps (Figure \ref{fig:sample_intro}), and that these instruction-action pairs can be easily augmented to include visual data, forming complete triads of vision-language-action for the training of autonomous vision-language navigation models. 

Recognizing that the range of information available from a cell phone application is more standardized and quantized than that of a human passenger (or backseat driver), in this research we analyze linguistic characteristics in readily-available automated verbal commands. The types of information available to the guided user can vary significantly across commands, which may prove sufficient or deficient towards particular VLA tasks \cite{roy2024doscenes}. Specifically, we are analyzing relationships between language and instruction characteristics as well as action recognition across several navigational applications to extract different combinations of IA pairs in response to similar scenarios. For example, two different navigation apps might describe the same maneuver in two different ways:

\begin{quote}
    \textit{“In a quarter mile, turn right onto East North Bear Creek Drive.”}
   
    \textit{“At the stop sign, turn right onto East North Bear Creek Drive.”}
\end{quote}

The first example makes a distance reference, while the second makes a reference to a static object (in this case, a road sign). Because statements may provide different referential cues, in this research we have categorized observed statements into the following attribute classes to illustrate the breadth of possible data generation:
\begin{enumerate}
    \item Road Names: Roadways referencing official names from rural streets, collectors streets, minor arterial, major arterial which can also be used as an indicator in navigation. 

    \item Distance of the Location: Separation within points measured in Feet and Miles.

    \item Static Objects: Objects in fixed positions serving a functional purpose integrated within the environment.

    \item Turn: An action of changing a vehicle position.

    \item Cardinal: An indication referring to a principal direction either being North, South, East, West.
    
    \item Location Name: Instruction that has a explicit reference to named geographical location, buildings, or desired destination
    
    \item Lane Information: Instructions regarding how to navigate incoming lane.
    
    \item Light Information: Instructions regarding how to navigate incoming lights.

\end{enumerate}
Prototypical examples of each of these class attributes are provided in Table \ref{tab:prototypes}, and attribute-tagged real-world example commands are provided in Table \ref{tab:specificexamples}.

Because of their underlying programmatic structure, navigational applications naturally create similar structured outputs, motivating our survey of the output range. When used as training data for VLN applications, small changes such as substituting ``quarter mile” for ``stop sign” can have different effects on model performance, relative to a robot's sensing and perception capabilities (e.g. ability to recognize visual features and/or ability to measure and understand distances \cite{cheng2025spatialrgpt}), leading to changes in how the instructions are executed or affecting the complexity of the decision-making.

We present two primary contributions in this research: 
\begin{enumerate}
    \item we explore three commonly available navigational applications to analyze variations in verbal command outputs towards utility in referential vision-language navigation models, and 
    \item we demonstrate an automated process to form annotated data consisting of video (vision), instruction (language), and vehicle trajectory (action), towards use as VLA model data. 
\end{enumerate}

\begin{table}
\caption{Sample Command Prototypes by Command Referentiality Class}

\centering
\begin{tabular}{|c|c|}
\hline
\textbf{Command Class} & \textbf{Example} \\ 
\hline
Distance        & ``Continue for half a mile."  \\
Turn              & ``Make a left turn." \\
Cardinal          & ``Head West."  \\
Road              & ``Turn onto Main Street."\\
Location Name     & ``Arrived at (location name)." \\
Lane Information  & ``Use the right two lanes."   \\
Light Information & ``Go past these lights."   \\
Static Object     & ``Go past the stop sign."  \\
\hline
\end{tabular}
\label{tab:prototypes}
\end{table}

\begin{table*}
\caption{Example Verbalized Command Outputs from Navigational Apps}
\centering
\begin{tabular}{|c|c|}
\hline
\textbf{Command Classification} & \textbf{Examples} \\ 
\hline
Distance, Turn, Road Name        & ``In 1000 feet turn left onto East 15th Street."  \\
Static Object, Lane Information, Road Name             & ``At the light use the left two lanes to turn left onto M Street Veterans boulevard." \\
Cardinal, Road Name           & ``Head West towards Lake Road, North Lake Road."  \\
Location Name              & ``Arrived at Pretty Good Burger."\\
Static Object, Light Information, Turn  & ``Go past these lights, and at the next set, turn left."   \\

\hline
\end{tabular}
\label{tab:specificexamples}
\end{table*}

\begin{figure*}
    \centering
    \includegraphics[width=\textwidth]{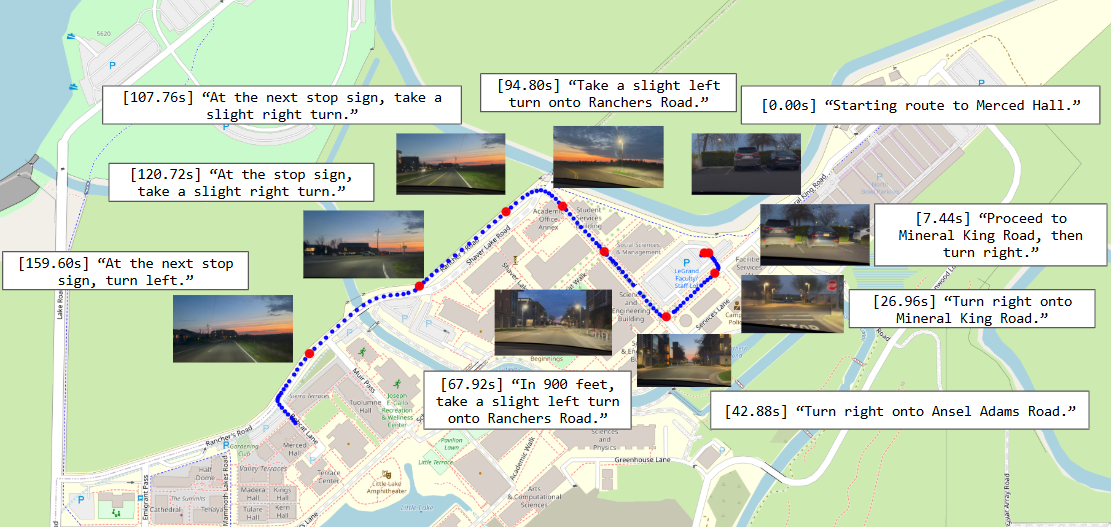}
    \caption{We demonstrated the capability of forming a fully automatic vision-language-action data generation system, a prototype of our proposed ADVLAT-Engine, using a single mobile phone. By logging GPS positions and recording video, there is sufficient information to synchronize frames to keypoints where GPS instructions are verbalized, extracted by a speech transcription module. Continuous frames between verbal events are available, making it possible for models to learn actions as sequences of video frame states or trajectory waypoints. As illustrated, language from the GPS system is instructive and scene-referential. In this demo, we synchronize video from an iPhone's native camera app, spatial positioning logs from the free myTracks app, and verbalized instructions from Apple Maps converted to text using OpenAI's whisper transcription model.}
    \label{fig:enter-label}
\end{figure*}

\section{Related Work}

This research presents a novel system for improved efficiency of data collection and curation for autonomous systems with applications in training safer and more effective self-driving systems. These systems require a significant amount of instruction-action data pairs, and our method provides a means to collect this volume while simultaneously automating the annotation process. In this section, we describe related research in automated data collection for safe autonomous driving tasks.

\subsection{Language Driven Navigation}

Language-driven navigation offers a bridge between human and computer interactions in the scene of autonomous vehicles and robotics. LM-Nav \cite{shah2023lm} uses natural language commands to guide autonomous vehicles in navigating real-world settings. They show that using natural language instructions and identifying key landmarks within these instructions can help create usable IA data for training. Subsequent models can use verbal commands referencing key landmarks with an image-language model and camera sensors to travel within an environment to locate key landmarks. 

As an example of the types of data that enable these systems to learn, the recently-released LeRobot dataset \cite{huggingfaceblog} provides multimodal sensor data from sixty vehicles in thirty German cities, paired with expert teacher and learning student driver actions guided by natural language instructions for a variety of driving tasks. Our research has strong similarity to this data scheme, with our command set left open to anything a navigational app would state, in comparison to LeNet's centrality to the EU driving license task standards. Further, our data pairs can be created automatically, without the need for an instructing teacher or copilot. 

\subsection{Vision-Language Datasets for Autonomous Vehicles}
Domain-specific Vision-Language-Action datasets are highly sought after in the AV and robotics field due to their scarcity, as many action sequences are constrained to the control mechanism of the autonomous system used to generate the data, enabling end-to-end learning and avoiding a learning stage to bridge the gap between abstract commands and actuated sequences. In recent years, there have been multiple high-quality dataset releases combining language and autonomous driving which exist in varying levels of abstraction, such as nuScenes-QA, doScenes, CoVLA, and Learning to Drive (L2D) \cite{qian2024nuscenes,roy2024doscenes, arai2024covla, huggingfaceblog}, paving the way for future vision-language data curators and providing a variety of modalities for developing a vision language model (VLM). These datasets are human-annotated, which has the benefit of high-quality natural language data at the expense of high (and sometimes intractable) levels of time and effort \cite{greer2024and}. Being able to automate this will solve a pressing issue in the creation process of these datasets and provide additional benefits such as improved cost efficiency and annotation consistency, preserving human effort for concentrated efforts in specific data domains for fine-tuning rather than routine and massive collection of highly similar data to satisfy foundational learning, an approach often taken with transfer learning or synthetic data \cite{casarin2024enhancing, jonker2024synthetic, mo2023few, wanggensim}.

\subsection{Automatic Data Annotation}

Recent advances in end-to-end learning and other data-driven solutions for autonomous vehicles has led to frameworks, algorithms, and multi-modal datasets to help automate annotations to meet this data volume demand. One example of automatic annotation is the Point Cloud Processing (PCP) algorithm of \cite{10588577}, where authors label 3D point cloud data from LiDAR and camera sensors with relevant object classes. Annotated point cloud data is valuable for 3D object detection, with these labels carrying implications for object meanings and object motion for agent path prediction and ego path planning. Further, the authors of \cite{10588577} incorporate their findings into LiDAR simulations in an autonomous driving simulator framework, used to further generate driving scenario data at the standards of OpenSCENARIO 2.0.


The essence of this research is its potential for automating the process of data annotation, reflecting analogous trends in direct use of internet-scale data to train foundation models in a variety of vision-related tasks. Even in semi-supervised approaches, such as OpenAnnotate2 \cite{zhou2024openannotate2}, 3D-BAT \cite{zimmer20193d}, and ActiveAnno3D \cite{ghita2024activeanno3d}, the ability to simplify the annotation workflow by performing complex multi-model annotation tasks with proposals or refinements from learned models helps to improve data annotation efficiency. We make an analogous improvement to the data pipeline for language-based command annotations rather than bounding box annotations, enabling an accelerated rate at which IA data is created.


\section{Methods}
\subsection{Pilot Data Collection}

For our development of a taxonomy of classes of referentiality available in GPS navigation commands, we collected a pilot dataset from a variety of navigational applications such as Google Maps, Apple Maps, and Waze, annotating common occurrences in the navigational systems shown in Table \ref{tab1}. Our pilot data included 71 commands from Apple Maps, 82 commands from Google Maps, and 80 commands from Waze, taken by driving 5 routes comprising residential, rural, and commercial areas, city streets, and freeways in California.

\begin{table*}
\caption{Statistics for Navigation Instruction References Based on Independently Collected Data }
\begin{center}
\begin{tabular}{|c|c|c|c|}
\hline
\textbf{Command Classification} & \textbf{Apple Maps} & \textbf{Google Maps} & \textbf{Waze Maps}\\
\hline
Turn               & 68  & 57 & 75\\
Road               & 49  & 77 & 61\\
Distance           & 22  & 45 & 45 \\
Static Object      & 41  & 19 & 3\\
Cardinal           & 2   & 8 
& 22 \\
Lane Information   & 0   & 3  & 3\\
Light Information  & 2   & 1  & 0\\
Location Name      & 1   & 0 
& 1\\
\hline
\end{tabular}
\label{tab1}
\end{center}
\end{table*}

\begin{table*}
\caption{Frequency of multi-attribute commands between the different navigation apps within the pilot dataset}
\begin{center}
\begin{tabular}{|c|c|c|c|c|}
\hline
\textbf{Command Classification} & \textbf{Google Maps} & \textbf{Apple Maps}  &\textbf{Waze Maps}
&\textbf{Total}
\\
\hline
Destination, Road, Turn & 21 & 20 & 37 & \textbf{78}\\
Road, Turn & 9 & 3 & 19 & \textbf{31}\\
Static Object, Turn &12 & 17 & 0 & \textbf{29} \\
Road, Static Object, Turn  & 5 & 23 & 0 & \textbf{28}  \\
Distance, Road & 19 & 0 &1 & \textbf{20} \\
Distance, Turn & 1 & 0 & 14 & \textbf{15} \\
Cardinal, Road & 3 & 1 & 0 & \textbf{4} \\
Distance, Lane, Road, Turn  & 2 & 0 & 1 & \textbf{3} \\
Static Object  & 0 & 0 & 3 & 
\textbf{3}\\
Distance, Destination & 2 & 0 & 1 & \textbf{3}\\
Distance, Lane, Turn & 0 & 0 & 2 & \textbf{2}\\
Cardinal, Turn, Road & 2 & 0 & 0 & \textbf{2} \\
Light, Static Object, Turn & 0 & 2 & 0 & \textbf{2}\\
Turn   & 0 & 1 & 1 & \textbf{2}\\
Cardinal, Distance, Road, Turn & 1 & 1 & 0 & \textbf{2}\\
Lane, Road, Turn & 1 & 0 & 0 & \textbf{1}\\
Lane, Road, Static Object & 1 & 0 & 0 & \textbf{1} \\

Distance, Destination, Turn & 1 & 0 & 0 & \textbf{1}\\

Destination, Location Name, Turn & 0 & 1 & 0 & \textbf{1} \\
Cardinal, Distance, Road &1 & 0 & 0 & \textbf{1} \\
Destination, Turn & 0 & 1 & 0 & \textbf{1}\\
Destination  & 0 & 0 & 1 & \textbf{1}\\

\hline
\end{tabular}
\label{tab_multi}
\end{center}
\end{table*}

Specifically, to collect data, we use a variety of navigational applications to track the turn-by-turn navigation system instructions while driving to specified destinations, producing a corpus of verbalized commands as data samples. We identified patterns within these verbal commands by manually observing referential patterns for subset categorization.
Table \ref{tab:prototypes} summarizes key differences in verbalized navigation instructions between Apple Maps, Google Maps, and Waze for the same routes. Table \ref{tab_multi} illustrates the frequency with which multiple command classes apply to a single instruction on each of the applications; for example, the combination of Destination, Road, and Turn appears most frequently. No instances were observed with more than four of our defined classes present. The majority of samples in the pilot data contained more than one class attribute.  


Comparing Apple Maps, Google Maps, and Waze Maps helps in the process of analyzing the structure of each application toward possible utility for particular VLA models or tasks, for cases when verbalized instructions with a particular style of referentiality may be most useful. 


Though the manual method of data collection via human annotation is time-consuming and lacks efficiency, our analysis of the range of commands motivates the use of this automated data modality toward VLA training, and we are able to completely replace human annotation with a fully-automated pipeline in the following sections, effectively removing the labor-based cost-limit to  the construction of instruction-action pairs to train models for language-driven autonomous driving.


\begin{figure*}
    \centering
    \includegraphics[width=0.89\textwidth]{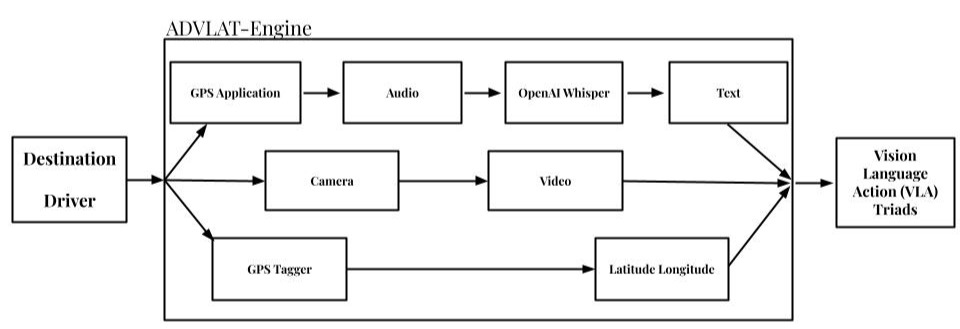} 
    \caption{System overview of the ADVLAT-Engine. The method used to collect data consisted of storing the text directions provided by the GPS, and recording the verbal instructions spoken aloud. These pairs are then manually annotated and sorted into various categories depending on the types on references made in the instruction, shown in the following tables.  }
    \label{fig:sample_advlat}
\end{figure*}

\subsection{Automated Data Collection System: Experimental Demonstration}

We designed a prototype automated GPS-app VLA data generator, which we refer to as \textbf{ADVLAT-Engine} (Autonomous Driving Vision-Language-Action Triad Engine), illustrated in Figure \ref{fig:sample_advlat}. The engine consists of the following simultaneous modalities: 
\begin{itemize}
    \item One or more visual perception streams,
    \item One or more language-based command streams, and 
    \item One or more action streams.
\end{itemize}

In our instantiation, we use a single outside-facing video camera for visual perception, recorded latitude and longitude as the action stream, and the audio output of the GPS app as the command stream. All modalities are synchronized, and future modifications can synchronize additional modalities such as LiDAR or CAN bus output to satisfy end-to-end learning requirements for particular VLN models.


\subsubsection{Transcription: Speech-to-text Audio Processing}
We perform audio transcription using the OpenAI whisper model \cite{radford2023robust}, providing timestamped text output from audio file segments. The speech recognition model is robust to spoken discourse from passengers as well as electronic synthesized or recorded output from the navigation apps, which can be recorded as physical device output or saved internally from a mobile device. Audio transcripts are particularly useful as event indicators while driving, since the GPS output indicates upcoming decisions, and can be used to identify notable action segments as opposed to 'stationary' driving periods (such as long stretches of freeway driving). Importantly, we view this as a means of data mining or data curation of novel or interesting scenarios, at least along the dimension of active action-taking by the driver, which can allow for a reduction of dataset size to only periods of action or control rather than overabundance of static or typical data, which is less beneficial to learning at high data volumes \cite{hacohen2022active}. 

We use off-the-shelf apps for recording timestamped video and GPS logs from mobile devices, allowing us to generate the sample IA data shown in Figure \ref{fig:enter-label}, demonstrating the generation of VLA triads by recording navigational instructions and logging GPS positions while recording road scenes simultaneously. As a result, we are able to link certain frames throughout road scenes to specific verbalized instructions, with a continuous flow of video and rover positions in between. The framework is extendable to human speech as well as automated application output. 





\section{Concluding Remarks: Applications \& Future Direction}

The efficiency in creating these datasets will prove beneficial in leveraging large volumes of data to develop and train vision-language-action models \cite{liu2024nvila}. Existing models which can learn bi-directional relationships between visual input and trajectories of vehicle positions provide a promising new direction for end-to-end learning for motion planning \cite{zhao2024drivellava, xing2025openemma}, and being able to boost the performance of these models with further training data at minimal cost could significantly progress the robustness, safety, and interactivity of autonomous navigation systems. 

It is the goal that these autonomous systems can, in the future, receive instructions from its passengers at the interval and scale of a GPS-style instruction \cite{moghadam2021autonomous}, linking the passenger's intention, driving scene, and navigational plan, such as ``pass this stop sign and drop me off at the bus stop." By automatically collecting and annotating data with information on key landmarks observed by visual sensors, models which can perceive and localize scene elements (e.g. the stop sign and bus stop) \cite{wei20203d} can effectively form trajectory motion plans which satisfy the instructed command. We propose that learning of motion on such time-scale may be achieved with constrastive methods, which have recently shown great success at foundation model training for language-based generation on a variety of tasks \cite{yang2022vision, goel2022cyclip, kopuklu2021driver}. The structured nature of a well-engineered automated annotation system can provide support for these contrastive learning approaches, ultimately boosting the robustness and accuracy of AV decision making, extending datasets for current perception and navigation foundation models to further driving domains and geographic regions. From large company fleets to single-vehicle research laboratories, many can find use in utilizing readily-generated, area-specific datasets as input to learning models to improve performance. Especially for self-driving vehicles, creating diverse datasets will decrease the performance gap found when comparing the accuracy of familiar environments against new environments.




\bibliographystyle{IEEEtran}
\bibliography{refs}

\end{document}